\def\year{2024}\relax
\documentclass[letterpaper]{article} % DO NOT CHANGE THIS
\usepackage{algorithm}
\usepackage{algorithmic}
\usepackage{newfloat}
\usepackage{listings}
\usepackage{aaai22}  % DO NOT CHANGE THIS
\usepackage{times}  % DO NOT CHANGE THIS
\usepackage{helvet}  % DO NOT CHANGE THIS
\usepackage{courier}  % DO NOT CHANGE THIS
\usepackage[hyphens]{url}  % DO NOT CHANGE THIS
\usepackage{graphicx} % DO NOT CHANGE THIS
\usepackage{natbib}  % DO NOT CHANGE THIS AND DO NOT ADD ANY OPTIONS TO IT
\usepackage{caption} % DO NOT CHANGE THIS AND DO NOT ADD ANY OPTIONS TO IT
\DeclareCaptionStyle{ruled}{labelfont=normalfont,labelsep=colon,strut=off} % DO NOT CHANGE THIS
\usepackage{algorithm}
\usepackage{algorithmic}
\usepackage{xcolor}

\usepackage{newfloat}
\usepackage{listings}
\floatstyle{ruled}
\newfloat{listing}{tb}{lst}{}
\floatname{listing}{Listing}
\usepackage{booktabs}

\title{Large Language Models Reveal Information Operation Goals, Tactics, and Narrative Frames}
\author{
    Keith Burghardt\textsuperscript{\rm 1}
    Kai Chen\textsuperscript{\rm 2}    
    Kristina Lerman \textsuperscript{\rm 1,2}    
}
\affiliations{
    \textsuperscript{\rm 1} USC Information Sciences Institute\\
    \textsuperscript{\rm 2} University of Southern California\\
   \{keithab,kchen,lerman\}@isi.edu 
}

\begin{document}

\maketitle

\begin{abstract}
Adversarial information operations can destabilize societies by undermining fair elections, manipulating public opinions on policies, and promoting scams. Despite their widespread occurrence and potential impacts, our understanding of influence campaigns is limited by manual analysis of messages and subjective interpretation of their observable behavior. In this paper, we explore whether these limitations can be mitigated with large language models (LLMs), using GPT-3.5 as a case-study for coordinated campaign annotation. We first use GPT-3.5 to scrutinize 126 identified information operations spanning over a decade. We utilize a number of metrics to quantify the close (if imperfect) agreement between LLM and ground truth descriptions. We next extract coordinated campaigns from two large multilingual datasets from X (formerly Twitter) that respectively discuss the 2022 French election and 2023 Balikaran Philippine-U.S. military exercise in 2023. For each coordinated campaign, we use GPT-3.5 to analyze posts related to a specific concern and extract goals, tactics, and narrative frames, both before and after critical events (such as the date of an election). While the GPT-3.5 sometimes disagrees with subjective interpretation, its ability to summarize and interpret demonstrates LLMs' potential to extract higher-order indicators from text to provide a more complete picture of the information campaigns compared to previous methods.
\end{abstract}

\section{Introduction}
Information operations (info-ops) often utilize social media platforms to manipulate users at scale~\cite{bradshaw2018global,badawy2018analyzing,kim2018uncover,burghardt2023socio}. Efforts include, but are not limited to: manipulating authentic users on each side of a divisive issue~\cite{ratkiewicz2011detecting,kim2018uncover,Stella2018}, reducing trust in democracies~\cite{badawy2019characterizing}, and promoting hate speech \cite{Hickey_Schmitz_Fessler_Smaldino_Muric_Burghardt_2023}, misinformation \cite{vosoughi2018spread}, and financial scams~\cite{Pacheco2021}. 

Info-ops utilize a range of techniques to accomplish their goals, often though the coordination of several accounts to promote a particular message~\cite{burghardt2023socio,Pacheco2021}, while simultaneously appearing to act organically, as a human user, so as to build trust and evade detection by social media platforms~\cite{Ferrara_2017,Sayyad2020,omsome2022_ukraine}. Despite substantial progress in detecting coordinated campaigns, whether via conventional methods \cite{luceri2023unmasking,burghardt2023socio,Pacheco2021} or even LLMs \cite{luceri2023leveraging}, there has so far been  little work in utilizing AI to analyze campaigns at scale \cite{burghardt2023socio}. %This is not to say 
Instead, analysis of campaigns has
%. Moreover, there are connections between clandestine attempts to influence and rhetorical frames political scientists utilize to understand opinion change \cite{chong2007framing}. Instead we lack the ability to capture at scale 
relied on manual review and annotation~\cite{martin2019trends,carley2020}, which reduces our ability to detect campaigns at scale and understand their goals and tactics. % the goals of campaigns that are especially harmful, let alone understand their tactics, which we can then attempt to mitigate.

In this paper, we explore the utility of LLMs at annotating campaign goals, tactics, and narrative frames, using GPT-3.5 as a case study. We utilize an annotated dataset of campaigns \cite{martin2019trends} as a benchmark to test the accuracy of zero-shot LLMs and find their outputs are substantially better than baselines, although they often make mistakes. We explore the utility of LLMs in previously under-studied campaigns using two multi-lingual X datasets from the 2022 French election and 2023 Balikatan Philippine-U.S. military exercise in 2023, respectively. These data contain several coordinated campaigns and discuss a range of concerns, including the Russia-Ukraine war, domestic politics, and the economy. We first extract previously unknown coordinated campaigns using proxies of coordination \cite{burghardt2023socio,luceri2023unmasking}, a recent but well-studied method that is both interpretable and achieves reasonable precision and recall across a range of ground truth datasets \cite{luceri2023unmasking}. This proxy links two accounts if they create posts with long strings of hashtags in the exact same order, thus generating a network of coordinated accounts where nodes are accounts and links are account pairs that are coordinated. We can then define campaigns as connected clusters of coordination. Within each coordinated campaign, we extract the concerns discussed in each post using a novel instruction-tuned LLama-2 model that achieves reasonable precision and recall. After filtering campaign posts by their respective concerns, we then apply GPT-3.5 to extract the validated features of prior campaigns \cite{martin2019trends}, such as the political goals, and countries responsible for the campaign. We further extract tactics via the BEND framework \cite{carley2020}, and frame narratives \cite{chong2007framing}, which is a political science theory used to understand how politicians frame topics to influence the populace. The LLM annotation demonstrates that framing offers a new understanding of information campaigns as well. 
To summarize, our contributions are:
\begin{itemize}
    \item We develop a novel method to validate LLM info-ops capabilities based on a large dataset of information operations \cite{martin2019trends}.
    \item We create a new concern detection model and apply it to two datasets in order extract issues pushed by info-ops.
    \item We utilize the validated LLM on dynamic and concern-tailored info-ops in order to evaluate their answers to various well-grounded questions about goals, tactics, and narrative frames.
\end{itemize}
Overall, we demonstrate LLMs offer rich high-level indicators than previously possible \cite{burghardt2023socio}, even if the tool is imperfect. While we do not expect LLMs to ever replace humans in analyzing information campaigns, we believe they can substantially speed up our understanding of each campaign.  Our code, including prompts and outputs, are shown in the following repository: \url{https://github.com/KeithBurghardt/LLM_Coordination}.

\section{Related Work}

\paragraph{Information Operations} %

Online manipulation \cite{tucker2018social} is both widespread and covers a range of goals such as politics (e.g., the Brexit vote~\cite{howard2016bots} and presidential elections~\cite{burghardt2023socio,Ferrara_2017,bessi2016,badawy2018analyzing,badawy2019characterizing,kim2018uncover}), psychological warfare \cite{omsome2022_ukraine}, promoting hate \cite{Hickey_Schmitz_Fessler_Smaldino_Muric_Burghardt_2023}, scams~\cite{Pacheco2021}, and pushing COVID-19-related messages \cite{Graham2020,PinaGarcia2022}. We cannot easily assess the impact of these information operations \cite{Freelon2020}, but their widespread use suggests they are useful for manipulation.

Notably, however, these analyses have been manual, either through exploring case studies (as above), or through a systematic categorization that have only been applied to a limited set of campaigns. For example, Carley developed a BEND framework \cite{carley2020} to categorize the tactics, such as Excitement of a topic or Distortion of a narrative, these categories have not been applied at scale. Similarly, 126 campaigns have been systematically studied \cite{martin2019trends} with their own set of categories, such as political goals, but it is unclear how well these methods extend to other campaigns. Finally, a completely different field, political science, has developed a parallel set of methods to understand narrative frames used to change the opinions of the populace, known as Framing Theory \cite{chong2007framing}, which aim to categorize the way campaigns frame problems, causes, and remedies, as well as how they are summarized via slogans. This theory has been used to understand social media campaigns \cite{hon2016social,shahid2020detecting}, but has been under-applied to understand information operations.

\paragraph{Detecting Information Operations}
There have been a range of tools to analyze information operations, starting with bot-detection tools often on X (formerly Twitter) \cite{Ferrara_2017,Stella2018}, and moving onto coordinated activity detection, namely whether accounts perform in concert to push a message \cite{starbird2019disinformation,burghardt2023socio}. These tools have been applied across multiple platforms (e.g., Facebook \cite{Giglietto2020,Giglietto2020Facebook} or YouTube \cite{kirdemir2022towards}, as well as Twitter \cite{Sharma2020,Nizzoli2021,Weber2021,Mazza2022,Cinelli2022,burghardt2023socio,luceri2023leveraging,luceri2023unmasking}). While a range of methods to detect account coordination are used \cite{Sharma2020,Weber2021,Schliebs2021,Pacheco2021,Schliebs2021,kirdemir2022towards,Giglietto2020Facebook,burghardt2023socio,Pacheco2021}, we stick to an interpretable and well-respected metric, hashtag co-occurence \cite{burghardt2023socio,luceri2023unmasking}, which has been independently verified as a useful feature to detect information operations \cite{luceri2023unmasking}. To find coordinated campaigns these data are converted into a network where accounts are nodes and links connect coordinated pairs \cite{Pacheco2021,burghardt2023socio}. Coordinated campaigns are then connected components within this network.

\paragraph{AI Analysis of Information Operations}
While LLMs have proven remarkably useful across a range of fields \cite{katz2024gpt,gpt_hsexam} (including analysis of social media \cite{luceri2023leveraging,liyanage2023gpt}, this field is still quite nascent, with papers just beginning to understand how LLMs could, e.g., detect coordination \cite{luceri2023leveraging}. While some previous work has used conventional methods to analyze stories within coordinated accounts \cite{Ehrett2021}, stances \cite{CHEN2021913}, or socio-linguistic indicators \cite{burghardt2023socio}, there is little work on LLMs to analyze information operations.

\begin{figure*}
    \centering
    \includegraphics[width=\linewidth]{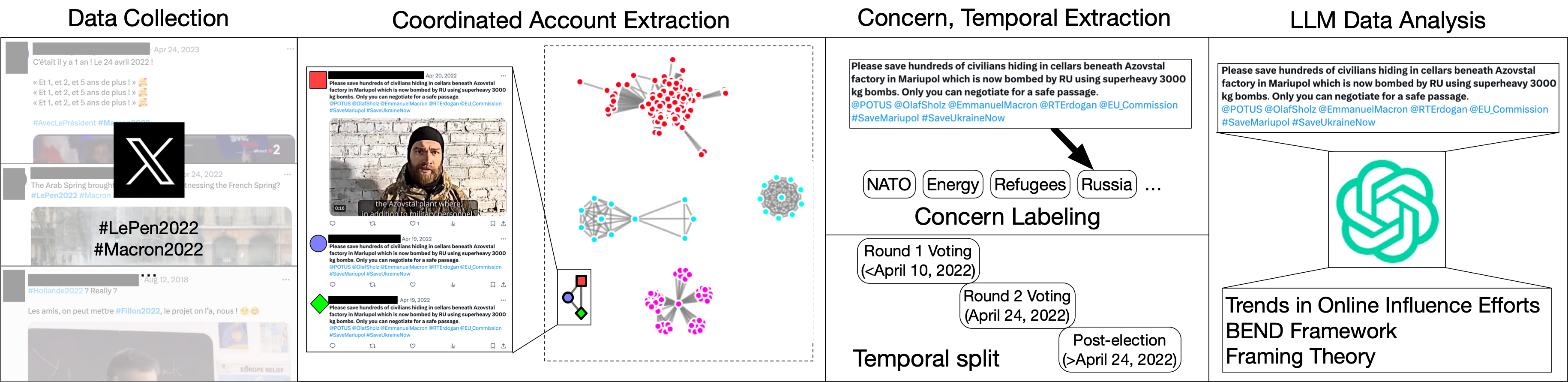}
    \caption{We develop an annotation technique as follows. First, we extract data from X matching a set of keywords related to the 2022 French election or the 2023 Balikatan  U.S.-Philippines military exercise. We then collect coordinated campaigns from these data based on a well-accepted hashtag co-occurrence metric \cite{burghardt2023socio,luceri2023unmasking}. We then label concerns and break up posts to particularly important date ranges for each campaign (such as just before an election). Alternatively, when testing the campaigns extracted in \cite{martin2019trends}, we use LLMs to create posts from these campaigns (we lack concern or dynamic information from the concern data). Finally, we utilize GPT to extract features based on previous framework \cite{martin2019trends}, the BEND framework \cite{carley2020}, and Framing theory \cite{chong2007framing}. We evaluate each output, to address the potential for halucinations, to understand the utility of the LLMs as methods to extract higher-order features of coordinated campaigns.}
    \label{fig:GPT-Schematic}
\end{figure*}

\paragraph{Concern detection.} 
Concerns, also known as wedge issues~\cite{van2014exploiting,heinkelmann2020divided}, is crucial for understanding social media influence campaigns~\cite{martinez2023examining}. Traditionally concern detection relies on open-world classification methods~\cite{shu2018unseen,esmaeilpour2022zero,bai2023open}, but these methods are limited by data size; moreover concern detection in social media is complicated and nuanced. \citet{burghardt2023socio} proposed a method that extracts concern keywords from Wikipedia for training models such as BERT~\cite{devlin2018bert}. This method is effective but depends on the Wikipedia pages and the quality of keyword extraction. 

Recent advances in LLMs have offered greater promise. For example, \citet{chen2024isamasred} utilize LLMs to enhance the extraction of keywords from Wikipedia pages. However, the high costs of powerful closed-source LLMs like GPT-4 \cite{achiam2023gpt} force a turn towards knowledge distillation. This approach transfers capabilities from larger to smaller models efficiently~\cite{gudibande2023false, chen2024susceptible}, an inspiration for our cost-effective concern detection framework, which does not utilize computationally costly LLMs and potentially expensive APIs. 
Finally, we utilize instruction tuning, which can enhance the performance of a tuned model, particularly in zero-shot tasks~\cite{mishra2021cross,sanh2021multitask,wei2021finetuned}.

\section{Methods}

Figure~\ref{fig:GPT-Schematic} shows the overall diagram of the proposed system. The components of the system are described in detail below.

\subsection{Data Collection}
We use Twitter data pertaining to the discussions of the 2022 French election (Round 1 was April 10, 2022 and Round 2 occurred on April 24, 2022), and the ``Balikatan'' U.S.-Philippines military exercise from April 11 to April 28, 2023. All data have post ID and PII are removed prior to analysis to improve anonymity and collection methods were approved by the respective institutional review boards. Data were publicly collected in the U.S. and did not require consent.

The 2022 French election dataset contains 5.9M posts (4.3M reposts, 1.6M replies, 18K original posts; 678K authors total) spanning February 15 to the end of June 30, 2022. Data were collected by a third party via keywords related to the French election, such as ``election'' or ``élection'' and presidential candidate accounts and their parties (@EmmanuelMacron, @ZemmourEric, or @MLP\_officiel). Within this dataset, 95\% of posts were in French and the rest (5\%) were in English based on X's language feature. 

In contrast, the U.S.-Philippines dataset contains 4.7M posts (3.2M reposts, 852 replies, 605K original posts, 130K quote posts; 1.9M authors) spanning January 1, 2023 to June 28, 2023. Data were similarly collected from a third party via keywords. Within this dataset, 94\% of posts were in English and the rest (6\%) were in Tagolog. 

\begin{figure}[tb]
    \centering
    \includegraphics[width=\linewidth]{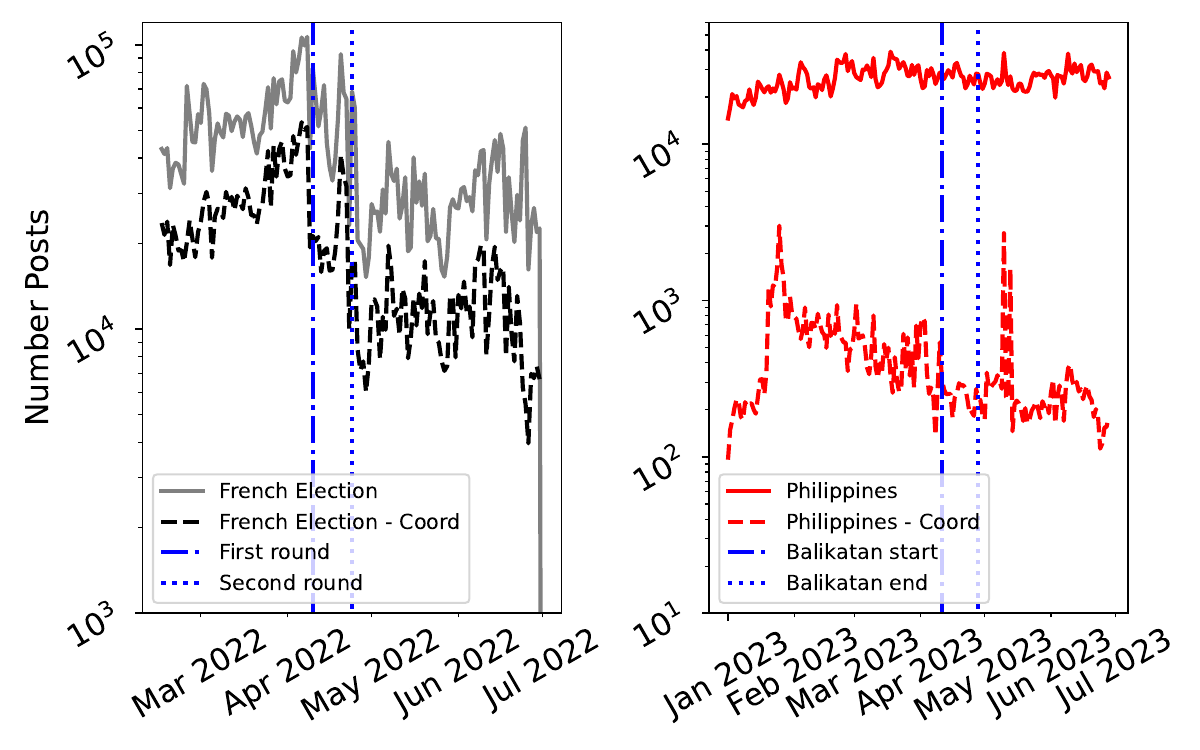}
    \caption{Number of posts over time from coordinated and non-coordinated accounts. (a) 2022 French election, with Round 1 and Round 2 elections labeled, and (b) 2023 Balikatan U.S.-Philippines military exercises that took place between April 11 and April 28, 2023.}
    \label{fig:post-freq}
\end{figure}

The LLM is therefore tested multilingual datasets, broadly in either French or English. We see posts over time in Fig.~\ref{fig:post-freq}, which shows the daily post count for coordinated and coordinated accounts (note the y-axis is log-scaled).

\subsection{Coordinated Account Extraction}
Inspired by \cite{burghardt2023socio,luceri2023unmasking,Pacheco2021}, we use the co-occurrence of hashtags in a given post as indicators of coordination, which has been well-validated in ground truth datasets \cite{luceri2023unmasking}. To this end, we detect whether two accounts are coordinated if they each contain posts with at least three hashtags, all of them the same and in the same order. This simple heuristic captures the notion that individual text may vary but if the text is semantically the same, usually the hashtags will be in the same order. This is independently validated on ground truth data and performs well with high precision and recall in many datasets \cite{luceri2023unmasking}. This indicator of coordination  creates networks whose accounts are linked if the indicator flags these accounts as coordinated. This in turn means we can define different coordination campaigns as separate connected components, which has been validated in previous work \cite{burghardt2023socio}. For coordination detection, we used Intel(R) Xeon(R) CPUs in an internal cluster.

\subsection{Concern Labeling}

Detecting concerns is a crucial component of identifying influence campaigns. A \emph{concern} refers to an issue characterized by multiple, competing, and often incompatible positions, commonly known as ``wedge issues''~\cite{van2014exploiting}. For instance, crime or environmental issues could be significant concerns to the Philippines' population. These concerns are rooted in deeper moral or sacred values, which play a pivotal role in shaping social identity and beliefs. By exploiting these concerns, adversarial actors can effectively widen social divisions or manipulate elections, leveraging the profound impact of moral and sacred values on societal dynamics \cite{ayala2010difference}.

Concern detection in social media is a complex and challenging task. Social media discussions touch on a broad array of concerns, many of which are dynamic and emerge from recent events. This variability requires methods that have a strong general ability to respond to new issues. Social media datasets are also massive and often lack sufficient context to accurately interpret and categorize concerns. This lack of context hampers the ability of models to make informed classifications. The boundaries between different concerns are often blurred, making it challenging to clearly delineate and classify them. This ambiguity complicates the task of concern detection and %requires sophisticated models that can handle such nuances. These challenges 
underscores the challenges of concern detection in social media.

\begin{figure*}[ht]
    \centering
    \includegraphics[width=1\linewidth]{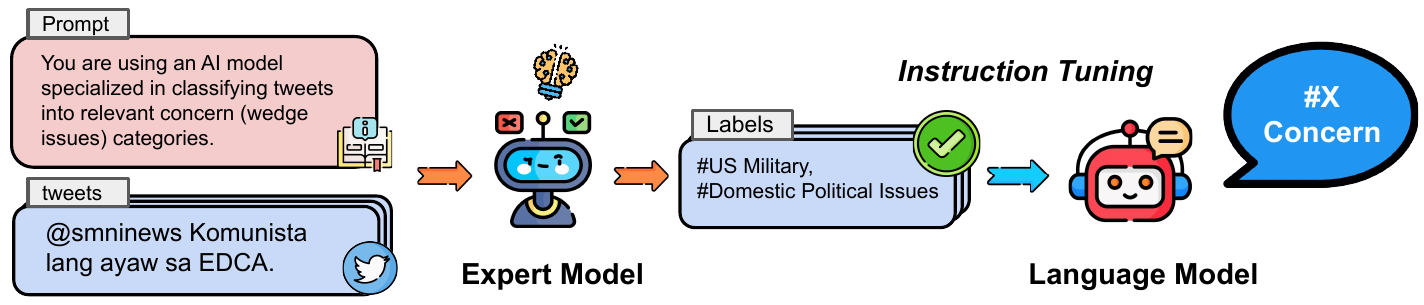}
    \caption{The framework of concern detection. Initially, we sample a small amount of data and label them using an expert model. Subsequently, we train a language model on this labeled dataset through instruction tuning.}
    \label{fig:concern-fram}
\end{figure*}

\begin{figure}
    \centering
    \includegraphics[width=\linewidth]{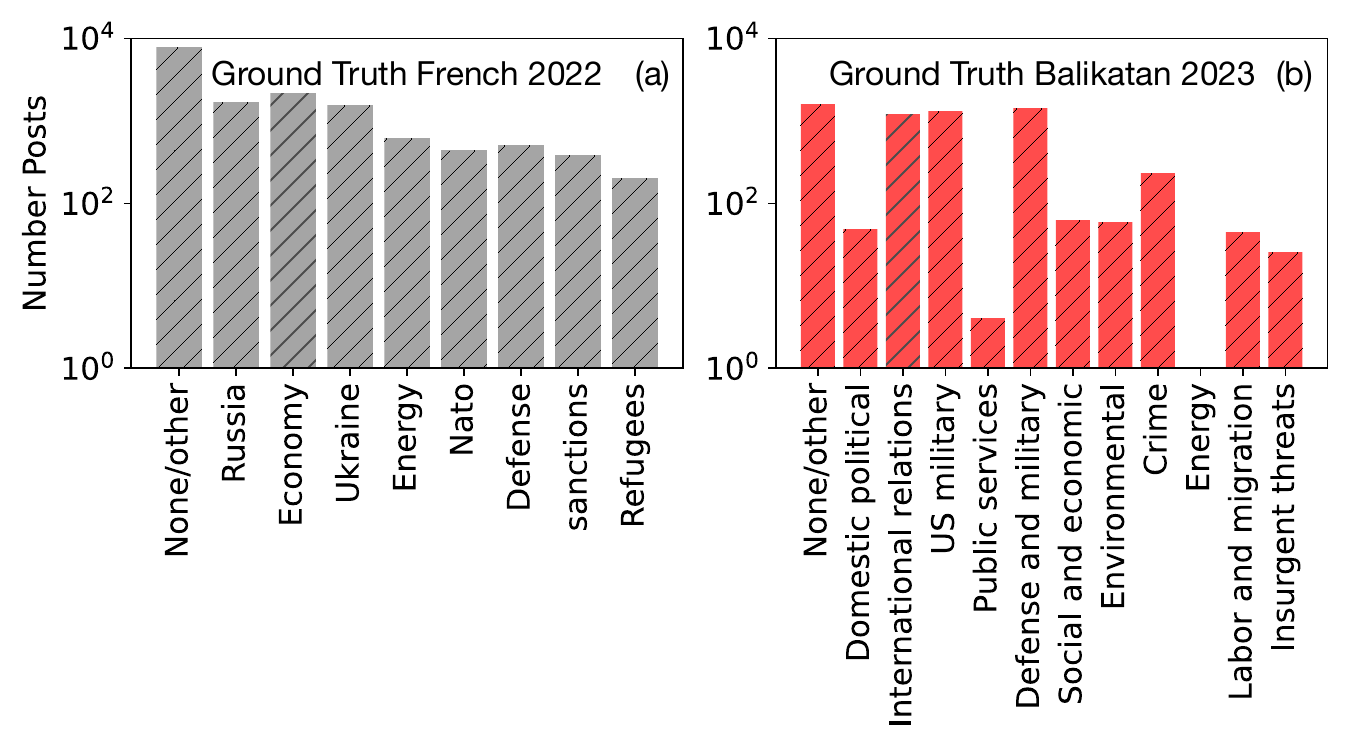}
        \caption{The frequency of concern posts in human annotated validation data.  (a) 2022 French election and (b) 2023 Balikatan U.S.-Philippines military exercises.}
    \label{fig:concern-gtfreq}
\end{figure}

To address these challenges, we propose a novel framework that leverages LLMs to automate concern detection. The overview of the framework is shown in Figure~\ref{fig:concern-fram}, and it is composed of the following steps:
\begin{itemize}
    \item \textbf{Sample classification.} Our first step is to annotate a sample of X posts. Given a set of concerns relevant to the domain identified by subject matter experts, we first create an \textit{expert model} for concern detection. The categories were wide-ranging, such as Russia, the economy, domestic politics, international relations, and the US military. The concerns are explained in more detail in the Results section. While there are other potential categories that could be identified, these nonetheless act as a proof-of-concept of our methodology.  We use GPT-4 \cite{achiam2023gpt} as the expert or teacher model as it has demonstrated remarkable ability to classify text in a zero-shot setting but is expensive to employ for the classification of 1M+ post datasets. More specifically, we use GPT-4 to classify 20K posts for each dataset (40K posts total) and 3.6-10K posts within each dataset are annotated by multiple subject matter experts for validation (10K posts for the 2022 French Election dataset, and 3.6K posts for the Balikatan datast; see Fig.~\ref{fig:concern-gtfreq} for the frequency of each concern, where each posts may have multiple concerns). The prompt for concern classification using GPT-4 is shown in Appendix \ref{sec:prompt_temp} (we allow GPT-4 to annotate multiple concerns for each post). While we do not validate GPT-4 itself, we believe it's results are valid due to its ability to generate training data for an accurate concern prediction model, as described below. %We use the expert model to classify the sampled posts based on predefined concerns. %Recently large language models like GPT-4 have demonstrated remarkable ability in zero-shot tasks and are also underpinned by their extensive training on a diverse array of social media datasets. Given this capability, we leverage GPT-4 for the classification of our sampled dataset. 
    
    \item \textbf{Instruction tuning.} We use a distilled student model (Llama-2-7B \cite{touvron2023llama}) to annotate millions of posts, which is trained via instruction tuning, a commonly used method to fine-tune LLM parameters to a specific task \cite{zhang2023instruction}. The specific format of the instruction pairs used for Llama-2 is shown in the Appendix \ref{sec:inst_temp}. For training, we use two NVIDIA A100 (80GB) GPUs for $3$ epochs, with batch size $1$, the gradient accumulation steps $16$, the learning rate $2\times10^{-5}$ (other details of the training can be seen in the link to our code).
\end{itemize}
We measure the model performance using precision, F1, and ROC-AUC.
%\textbf{Experiment setup.} In sample labeling, we leverage GPT-4 to label twenty thousand social media data. We finetune Llama-2-7B using instruction tuning  with two NVIDIA A100 (80GB) GPUs for $3$ epochs, with batch size $1$, the gradient accumulation steps $16$, the learning rate $2e-5$.

\subsection{LLM Information Campaign Annotation}

We use three frameworks for LLM annotation with GPT-3.5 (using the version as of March 12-21, 2024).  While other models (including GPT-4) might be more accurate, they are too expensive for the broad range of questions we ask and tweets we have to annotate (the GPT-3.5 annotations in this paper cost \$50 and GPT-4 would, at present, swell that number 20-fold). There are also ways to further improve the prompts via chain-of-thought prompting or few-shot learning, but we use zero-shot prompting as a starting point.

We first annotated campaigns based on categories developed from \cite{martin2019trends}: 
\begin{itemize}
    \item Targeted country
    \item Attacking country
    \item Political goal (category and description)
    \item Information operation description
\end{itemize}
Because we have a set of 126 annotated campaigns \cite{martin2019trends}, we can directly compare LLM annotations to ground truth descriptions in order to validate our zero-shot prompts.
%These are both used to help validate the LLM and can be used in the future to see if posts are part of, or consistent with, past information operations.

Next, we create prompts from the BEND framework \cite{carley2020}. Because ``B'' and ``N'' objectives (e.g., actions that decrease the importance of an opinion leader) represent actions that cannot be measured from posts alone, we focus instead on the ``E'' (positive) and ``D'' (negative) objectives:
\begin{itemize}
    \item Engage: bring up related/relevant topics
    \item Explain: provide details on or elaborate the topic
    \item Excite: elicit positive emotions
    \item Enhance: encourage messages on the topic
    \item Dismiss: explain why topic is unimportant
    \item Distort: alter the main message of a topic
    \item Dismay: elicit negative emotions
    \item Distract: discuss a different, irrelevant topic
\end{itemize}
Third, we borrow from Framing Theory \cite{chong2007framing,entman1993framing}, a well-studied problem in political science, to understand how issues are framed to influence the public. While typically applied to news media or politicians, we show these features can help us understand information operations as well. To this end, we prompt the LLM to provide:
\begin{itemize}
    \item The problem the information operation is describing
    \item The cause of the problem
    \item The remedy for this cause
    \item Metaphors used in this frame
    \item The catchphrases and slogans
    \item Frame examples for the  problem, cause, and remedy (for qualitative validation)
\end{itemize}

In addition, we asked the LLM to provide:
\begin{itemize}
    \item Cultural cues or in-group language
    \item Textual motifs (which was discarded due to irrelevant responses)
    \item Intentional rhetorical fallacies
\end{itemize} 
For the last bullet point, we included a list of intentional rhetorical fallacies in the prompt (e.g., Ad Hominem, Ad Baculum, etc.) from the Internet Encyclopedia of Philosophy (\url{https://iep.utm.edu/fallacy/}). While we cannot verify the accuracy of the LLM for the latter two sets of tasks (as we do not have validation data) the qualitative evaluation offers an important exploration of LLM annotation quality to guide future understanding of coordinated accounts.

\subsection{Evaluating LLM Coordination Annotation}
LLMs often suffer from hallucinations \cite{yao2023llm}, thus we want to evaluate how well LLMs can accurately explain the goals of information campaigns. To this end, we extract from prior work \cite{martin2019trends} descriptions of 126 information operations. Our goal is to check if LLMs could create semantically equivalent statements about these operations across various categories. 

We therefore fed into GPT-3.5 a prompt consisting of all information about the info-ops provided by the prior dataset \cite{martin2019trends}, namely the targeted country, the attacking country, the	political goal category, the political goal, and the event description. We then asked that the LLM generate 10 posts from X. After creating these posts, we then create a prompt for the LLM to determine the targeted country, the attacking country, political goal (category and description) and a description of the event. We finally compare the output of this model against the ground truth description.  Because we use the ground truth to generate the data (we lack the actual posts from the information campaigns described in \cite{martin2019trends}), there is some data leakage, thus our results act as an upper bounds to zero-shot GPT-3.5 performance. That said, the results are likely ecologically valid as we expect LLMs are being increasingly used for information operations. Example posts are in the Appendix Table~\ref{tab:LLm_ex}. While some of these X-like posts seem too on-the-nose (e.g. \#DiscreditKurz), the posts also qualitatively resemble posts from real coordinated accounts (in part because LLMs may be used in many campaigns).

LLM outputs can often be evaluated via simple metrics, such as F1 or ROC-AUC, as seen in the concern detection LLM model above, among others \cite{chang2023survey}. Other metrics might inclue BLEU or METEOR scores when evaluating LLMs' ability to translate text \cite{banerjee2005meteor}, but these score n-gram similarity rather than semantic similarity, which we are interested in. We therefore utilize NLI techniques \cite{chang2023survey} to determine if the LLM output and the ground truth description are equivalent. We use BART \cite{lewis2019bart} fine-tuned for natural language inference (\url{https://huggingface.co/facebook/bart-large-mnli}), GPT-3.5, and GPT-4 to determine whether the GPT-3.5 LLM output and the ground truth text are semantically equivalent. BART is given a hypothesis (the ground truth campaign description) and determines the confidence that this falls out from the premise, the LLM output. GPT-3.5 and GPT-4 are each prompted to determine whether the ground truth description and LLM output are equivalent (see link to our code for prompr details). While each metric may have its flaws \cite{tao2023eveval}, the consistency of all metrics improves the robustness findings (the prompts are in our code listed in the Introduction). %After determining whether any descriptions are equivalent, we label values "1" if they are and "0" if they are not. We then can measure accuracy as the number of times the statements are equivalent. Finally, country codes in the ground truth data were often misunderstood by these models, so we converted countries targeted or attacking to a standard naming convention. We use the majority class (most common ground-truth description) as a baseline.

\begin{table*}[htb!]
\centering
\scriptsize
\caption{2022 French Elections: Performance of Concern Labeling}
\begin{tabular}{l|ll|ll|ll}
\hline
Performance metric & \multicolumn{2}{c|}{Precision} & \multicolumn{2}{c|}{F1} & \multicolumn{2}{c}{ROC-AUC} \\
\hline
 & Model & Baseline & Model & Baseline & Model & Baseline \\
\hline
None or Other & 0.828$\pm$0.005 & \textbf{0.878$\pm$0.003}& \textbf{0.806$\pm$0.004} & 0.782$\pm$0.004 & \textbf{0.789$\pm$0.005} & 0.5\\
Russia & \textbf{0.906$\pm$0.004} & 0.289$\pm$0.006 & \textbf{0.846$\pm$0.004} & 0.169$\pm$0.004 & \textbf{0.808$\pm$0.005} & 0.5\\
Economy & \textbf{0.829$\pm$0.006} & 0.356$\pm$0.006 & \textbf{0.689$\pm$0.006} & 0.217$\pm$0.004 & \textbf{0.656$\pm$0.005} & 0.5\\
Ukraine & \textbf{0.944$\pm$0.003} & 0.289$\pm$0.006 & \textbf{0.830$\pm$0.005} & 0.169$\pm$0.004 & \textbf{0.773$\pm$0.006} & 0.5\\
Energy & \textbf{0.88$\pm$0.01} & 0.117$\pm$0.004 & \textbf{0.79$\pm$0.01} & 0.062$\pm$0.003 & \textbf{0.73$\pm$0.01} & 0.5\\
NATO & \textbf{0.87$\pm$0.01} & 0.084$\pm$0.004 & \textbf{0.74$\pm$0.01} & 0.044$\pm$0.002 & \textbf{0.68$\pm$0.01} & 0.5\\
Defense & \textbf{0.65$\pm$0.02} & 0.096$\pm$0.004 & \textbf{0.59$\pm$0.01} & 0.051$\pm$0.002 & \textbf{0.567$\pm$0.008} & 0.5\\
Economic Sanctions & \textbf{0.92$\pm$0.03} & 0.074$\pm$0.004 & \textbf{0.55$\pm$0.01} & 0.039$\pm$0.002 & \textbf{0.530$\pm$0.006} &0.5\\
Refugees & \textbf{0.72$\pm$0.03} & 0.04$\pm$0.003 & \textbf{0.61$\pm$0.02} & 0.02$\pm$0.001 & \textbf{0.57$\pm$0.01} & 0.5\\\hline
\end{tabular}
\label{tab:french}
\end{table*}

\section{Results}
\subsection{Evaluating Concern Labels}

To validate the reliability and practicality of concern detection, we conduct two validation experiments focusing on posts collected for distinct geopolitical domains: the 2022 French elections  and the Philippines. Both datasets are multilingual. %and are comprised of multiple social media data, including twitter and Reddit. 
In the context of the 2022 French elections, the analysis is structured around nine predefined concern categories: \emph{Ukraine, Russia, NATO, refugees, defense, economy, economic sanctions, energy}, and a residual category designated as \emph{none or other} to encapsulate unclassified concerns.
The Philippines Balikatan data involves twelve concerns: \emph{crime, defense and military, domestic political issues, international relations, labor and migration, public services, social and economic issues, US military, energy}, and a similarly defined \emph{none or other} category.
These categories were selected to cover a broad range of social and political concerns, allowing for an evaluation of the framework's capacity to navigate complex thematic content across disparate linguistic and cultural narratives.
%In each experiment, we employ GPT-4 to annotate a subset of twenty thousand posts from the respective datasets. Subsequently, these labeled posts are used to fine-tune the Llama2 model, adapting it to the specific concerns of each geopolitical context.
%For the 2022 French Elections dataset, the performance of the fine-tuned Llama2 model is evaluated against a ground truth dataset comprising ten thousand annotated posts. Similarly, in the Balikatan concern detection experiment, the model's efficacy is assessed using approximately six thousand annotated posts.
%The ground truth for both experiments were annotated by a panel of third party annotators, ensuring the accuracy and reliability of the evaluation process. 
%We measure the model performance using precision, F1, and ROC-AUC.

Table~\ref{tab:french} reports performance of the concern detection model on the French Elections dataset, where standard deviations reported are from bootstrapping the prediction and ground truth pairs 1000 times (due to the long training time of several hours, bootstrapping the training data and retraining the models was not feasible). The baselines were predicting all the same label (e.g., all posts are ``economy'', ``none or other'', etc. The model has relatively high average precision and F1 score, with especially strong performance for concerns like \emph{Ukraine}, \emph{Russia}, and \emph{NATO}. This suggests the model is effectively identifying relevant concerns with few false positives. However, the \textit{Refugees} and \textit{Defense} categories exhibit weaker performance, indicating potential difficulties in capturing the nuance within these concerns due in part to their low frequency in posts. %The ROC-AUC values are lower compared to the precision and F1 scores across all categories, which could point to a less favorable trade-off between precision and recall at various threshold settings.
Table~\ref{tab:balikatan} reports performance of the model on the Balikatan dataset.  As before,  we use the same baseline and standard deviations reported are from bootstrapping the prediction and ground truth pairs 1000 times. The \textit{Defence and military} and \textit{Public services} concerns demonstrate lower performance in part due to the low frequency of these events.  It is less challenging for the model to recognize concerns like \textit{Crime} and \textit{Labor and migration}. Energy has only 1 annotated post, so we do not validate it.

%Performance evaluation on the two datasets. The performance of the identified concerns is assessed individually using precision, F1 score, and ROC-AUC metrics.

\begin{table*}[htb!]
\centering
\scriptsize
\caption{2023 Balakitan Military Exercise: Performance of Concern Labeling}
\begin{tabular}{l|ll|ll|ll}
\hline
Performance metric & \multicolumn{2}{c|}{Precision} & \multicolumn{2}{c|}{F1} & \multicolumn{2}{c}{ROC-AUC} \\
\hline
 & Model & Baseline & Model & Baseline & Model & Baseline \\
\hline
None or other&\textbf{0.742$\pm$0.02}&0.533$\pm$0.016&\textbf{0.72$\pm$0.016}&0.696$\pm$0.014&\textbf{0.71$\pm$0.014}&0.5\\
Domestic political issues&\textbf{0.163$\pm$0.027}&0.048$\pm$0.007&\textbf{0.259$\pm$0.037}&0.092$\pm$0.013&\textbf{0.739$\pm$0.035}&0.5\\
International relations&\textbf{0.305$\pm$0.031}&0.106$\pm$0.01&\textbf{0.414$\pm$0.034}&0.191$\pm$0.016&\textbf{0.735$\pm$0.025}&0.5\\
US military&\textbf{0.448$\pm$0.026}&0.216$\pm$0.013&\textbf{0.565$\pm$0.025}&0.355$\pm$0.017&\textbf{0.754$\pm$0.016}&0.5\\
Public services&0.0$\pm$0.0&\textbf{0.004$\pm$0.002}&0.0$\pm$0.0&\textbf{0.008$\pm$0.004}&0.492$\pm$0.002&\textbf{0.5}\\
Defense and military&\textbf{0.637$\pm$0.052}&0.252$\pm$0.014&0.329$\pm$0.033&\textbf{0.402$\pm$0.018}&\textbf{0.59$\pm$0.014}&0.5\\
Social and economic issues&\textbf{0.285$\pm$0.081}&0.033$\pm$0.005&\textbf{0.277$\pm$0.073}&0.064$\pm$0.01&\textbf{0.626$\pm$0.04}&0.5\\
Environmental issues&\textbf{0.794$\pm$0.097}&0.04$\pm$0.006&\textbf{0.506$\pm$0.081}&0.077$\pm$0.012&\textbf{0.686$\pm$0.039}&0.5\\
Crime&\textbf{0.746$\pm$0.09}&0.021$\pm$0.005&\textbf{0.795$\pm$0.069}&0.041$\pm$0.009&\textbf{0.926$\pm$0.038}&0.5\\
Energy&---&---&---&---&---&---\\
Labor and migration&\textbf{0.604$\pm$0.102}&0.025$\pm$0.005&\textbf{0.573$\pm$0.088}&0.049$\pm$0.009&\textbf{0.773$\pm$0.052}&0.5\\
Insurgent threats&\textbf{0.105$\pm$0.107}&0.02$\pm$0.004&\textbf{0.063$\pm$0.062}&0.039$\pm$0.008&\textbf{0.52$\pm$0.024}&0.5\\
%\hline
%None or Other&	\textbf{0.621$\pm$0.01}&0.613$\pm$0.008&	\textbf{0.641$\pm$0.013}&0.442$\pm$0.008&	\textbf{0.668$\pm$0.008}&0.5\\
%Domestic political issues&	\textbf{0.054$\pm$0.009}&0.026$\pm$0.004&	\textbf{0.028$\pm$0.005}&0.013$\pm$0.002&	\textbf{0.704$\pm$0.031}&0.5\\
%International relations&	\textbf{0.528$\pm$0.011}&0.495$\pm$0.009&	\textbf{0.456$\pm$0.012}&0.329$\pm$0.008&	\textbf{0.63$\pm$0.008}&0.5\\
%US Military&	\textbf{0.582$\pm$0.011}&0.525$\pm$0.009&	\textbf{0.517$\pm$0.012}&0.356$\pm$0.008&	\textbf{0.66$\pm$0.008}&0.5\\
%Public services&	0.0$\pm$0.0&\textbf{0.002$\pm$0.001}&	0.0$\pm$0.0&\textbf{0.001$\pm$0.001}&	0.493$\pm$0.001&\textbf{0.5}\\
%Defense and Military&	0.359$\pm$0.014&\textbf{0.566$\pm$0.008}&	\textbf{0.792$\pm$0.019}&0.395$\pm$0.008&	\textbf{0.596$\pm$0.006}&0.5\\
%Social and Economic &	\textbf{0.069$\pm$0.023}&0.034$\pm$0.004&	\textbf{0.047$\pm$0.016}&0.017$\pm$0.002&	\textbf{0.542$\pm$0.022}&0.5\\
%Environmental Issues&	\textbf{0.451$\pm$0.059}&0.032$\pm$0.004&	\textbf{0.48$\pm$0.069}&0.016$\pm$0.002&	\textbf{0.711$\pm$0.032}&0.5\\
%Crime&	\textbf{0.947$\pm$0.011}&0.121$\pm$0.007&	\textbf{0.937$\pm$0.016}&0.064$\pm$0.004&	\textbf{0.976$\pm$0.007}&0.5\\
%Energy&	0.0$\pm$0.0&\textbf{0.001$\pm$0.001}&	\textbf{0.0$\pm$0.0}&\textbf{0.0$\pm$0.0}&	0.499$\pm$0.0&\textbf{0.5}\\
%Labor and Migration&	\textbf{0.184$\pm$0.042}&0.024$\pm$0.004&	\textbf{0.127$\pm$0.031}&0.012$\pm$0.002&	\textbf{0.657$\pm$0.037}&0.5\\
%Insurgent Threats&	\textbf{0.181$\pm$0.062}&0.014$\pm$0.003&	\textbf{0.136$\pm$0.05}&0.007$\pm$0.001&	\textbf{0.632$\pm$0.047}&0.5\\
\hline
\end{tabular}
\label{tab:balikatan}
\end{table*}

\subsection{LLM Evaluation}
\begin{figure}
    \centering
    \includegraphics[width=0.9\linewidth]{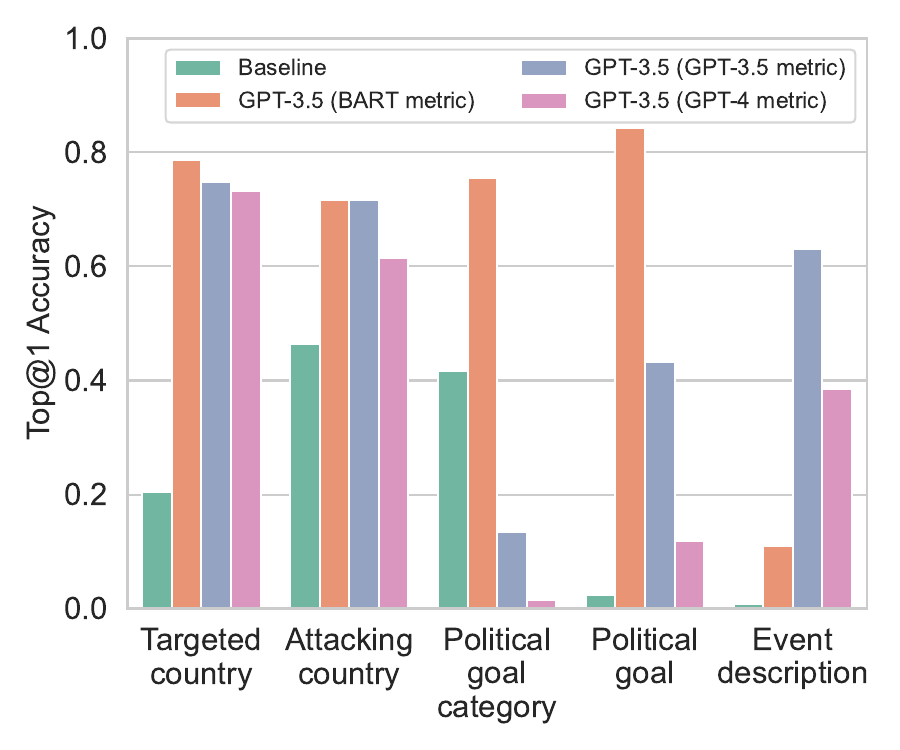}
    \caption{LLM metrics comparing GPT-3.5 annotations and ground truth from \cite{martin2019trends}. We use BART, GPT-3.5, and GPT-4 to evaluate whether GPT-3.5-based descriptions of posts agree with ground-truth data. We find across all metrics typically strong agreement with ground-truth, except for zero-shot political category.}
    \label{fig:validation}
\end{figure}

Our evaluation is shown in Fig.~\ref{fig:validation}. We find that the accuracy (how many descriptions are equivalent) far exceed baselines, especially for event descriptions, political goals, and targeted countries. Therefore while the LLMs are far from perfect, they can reasonably accurately infer many subtle details of a campaign. The performance is poorest, however, for the political goal category. We decided for consistency we would not present the set of possible categories to the LLM, which means the performance is necessarily poor when the LLM does not know the categories a priori. In contrast, not providing the ``country'' categories in advance did not substantially harm the LLMs ability to detect the targeted and attacking country. 

\subsection{Discovering Coordinated Accounts}
\begin{figure}
    \centering
    \includegraphics[width=\linewidth]{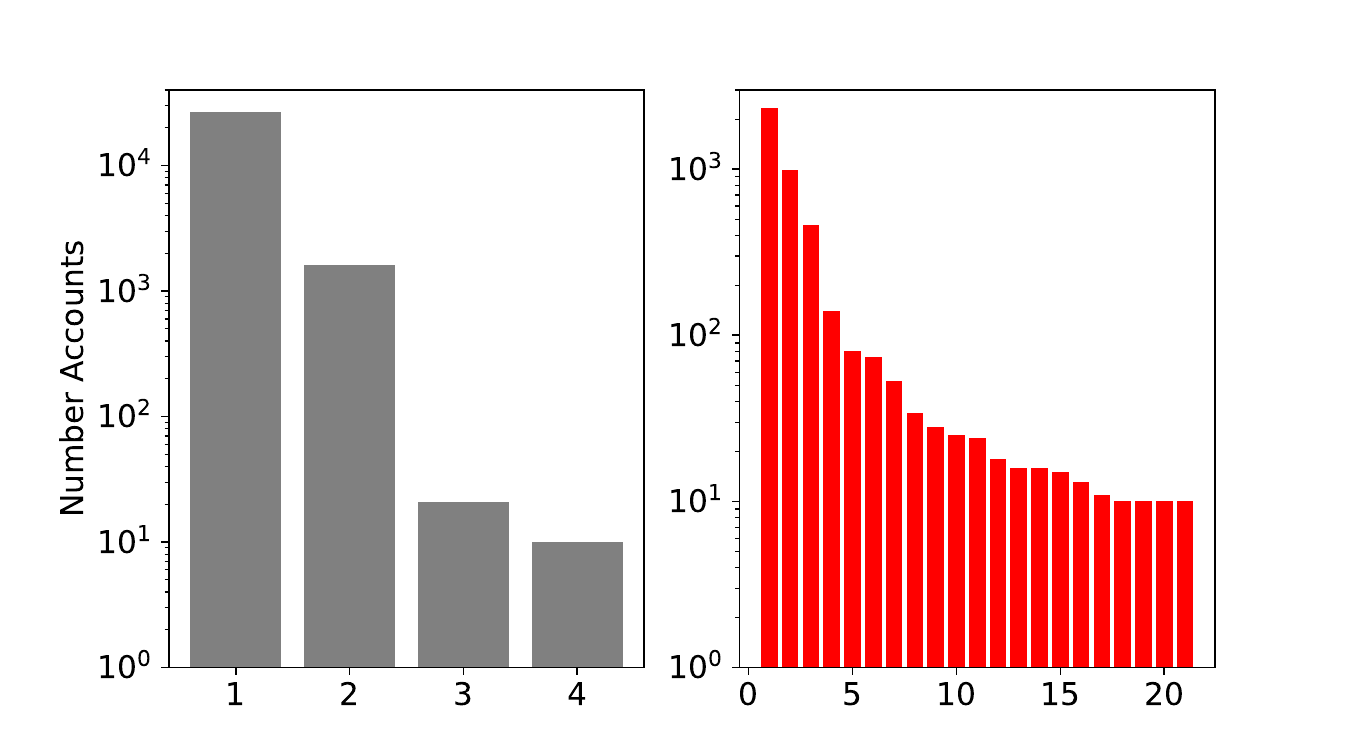}
    \caption{The number of coordinated accounts in each connected component (ranked by size).  Accounts link with each other if they share the same long string of hashtags in each post. (a) 2022 French election and (b) 2023 Balikatan U.S.-Philippines military exercises.}
    \label{fig:coord_freq}
\end{figure}

\begin{figure*}
    \centering
    \includegraphics[width=0.85\linewidth]{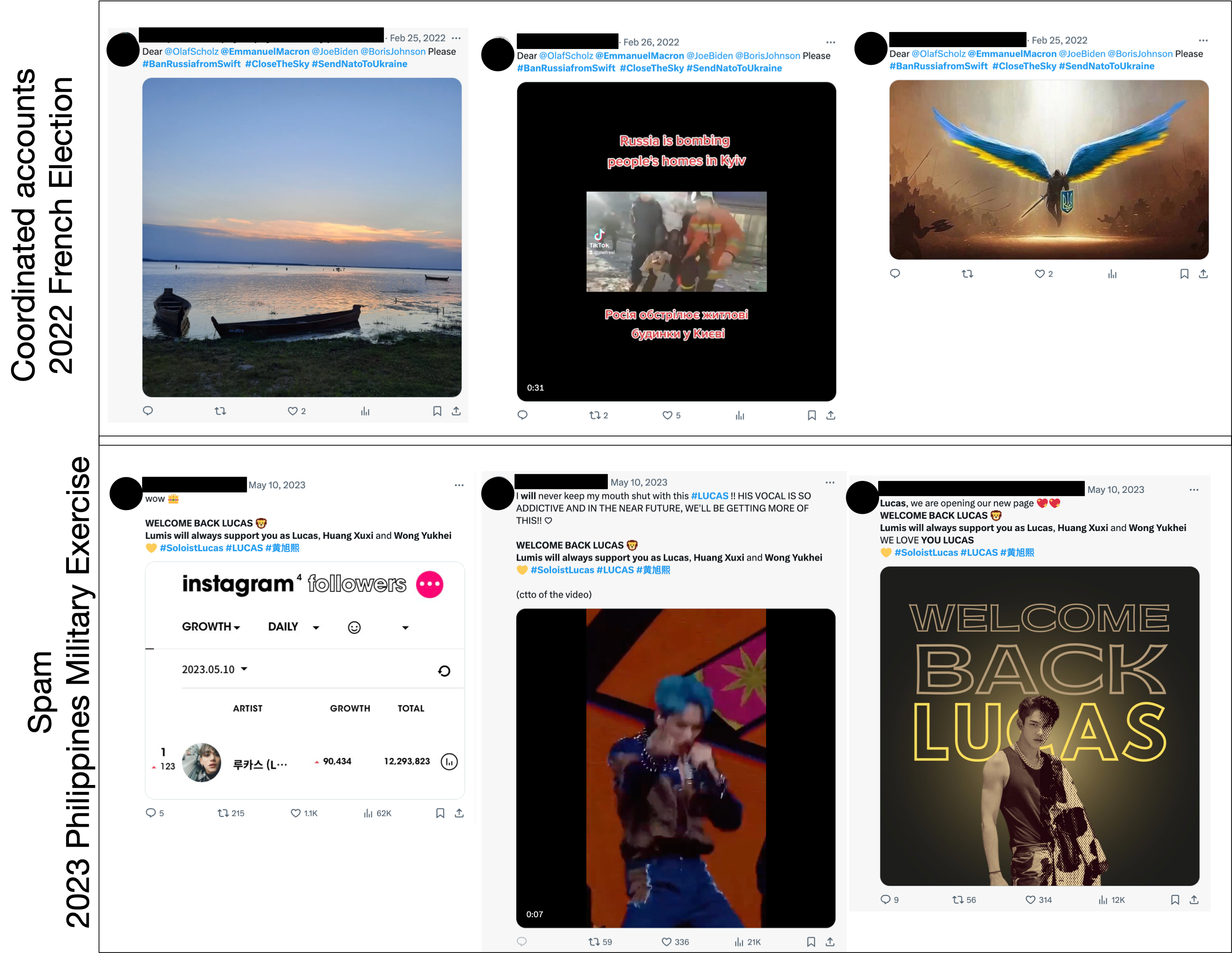}
    \caption{Example coordinated account posts for (top) the 2022 French election and (bottom) the 2023 Philippines Balikatan dataset.}
    \label{fig:examples}
\end{figure*}

We use coordination indicators to identify clusters of coordinated accounts in our French elections and Balikatan  datasets. We show the number of coordinated accounts in each cluster in Fig.~\ref{fig:coord_freq}. In the plots, we limit to clusters with at least 10 accounts each. In total, there were 29K coordinated accounts in the 2022 French election dataset, and 4.7K coordinated accounts in the 2023 Balikatan dataset. This corresponds to 2.7M posts in the 2022 French election (an astonishing 45.5\% of all posts) and 79K in the Balikatan dataset (1.7\% of all posts). While the former figure sounds surprising, they were written by merely 11.6\% of all accounts (0.2\% of accounts were coordinated in the Balikatan dataset). That said, 11.6\% is still a somewhat high figure and may reflect the keywords and accounts captured in the data sample. Another peculiarity of the French election dataset is that there are just 4 clusters with more than 10 accounts. The largest cluster corresponds to the largest proportion of all accounts (26.9K or 94\% of all coordinated accounts). Examples of the coordinated clusters are shown in Fig.~\ref{fig:examples}. In this figure, which shows the second largest cluster in the 2022 French election (top) and the largest cluster in the 2023 Balikatan dataset, we see examples of distinct accounts sharing near-identical posts, which are strong indicators of inauthentic online activity. 

Interestingly, while the coordinated accounts from the 2022 French election were posts apparently directed towards Macron and the French public, we see many coordinated accounts unrelated to Balikatan within the latter dataset. This may represent a lack of information operations targeting the Balikatan event or that information operations were using language we did not capture in the  keywords used to collect the posts, which is also represented by the relative lack of posts reposted or written by coordinated accounts.

\begin{figure}
    \centering
    \includegraphics[width=\linewidth]{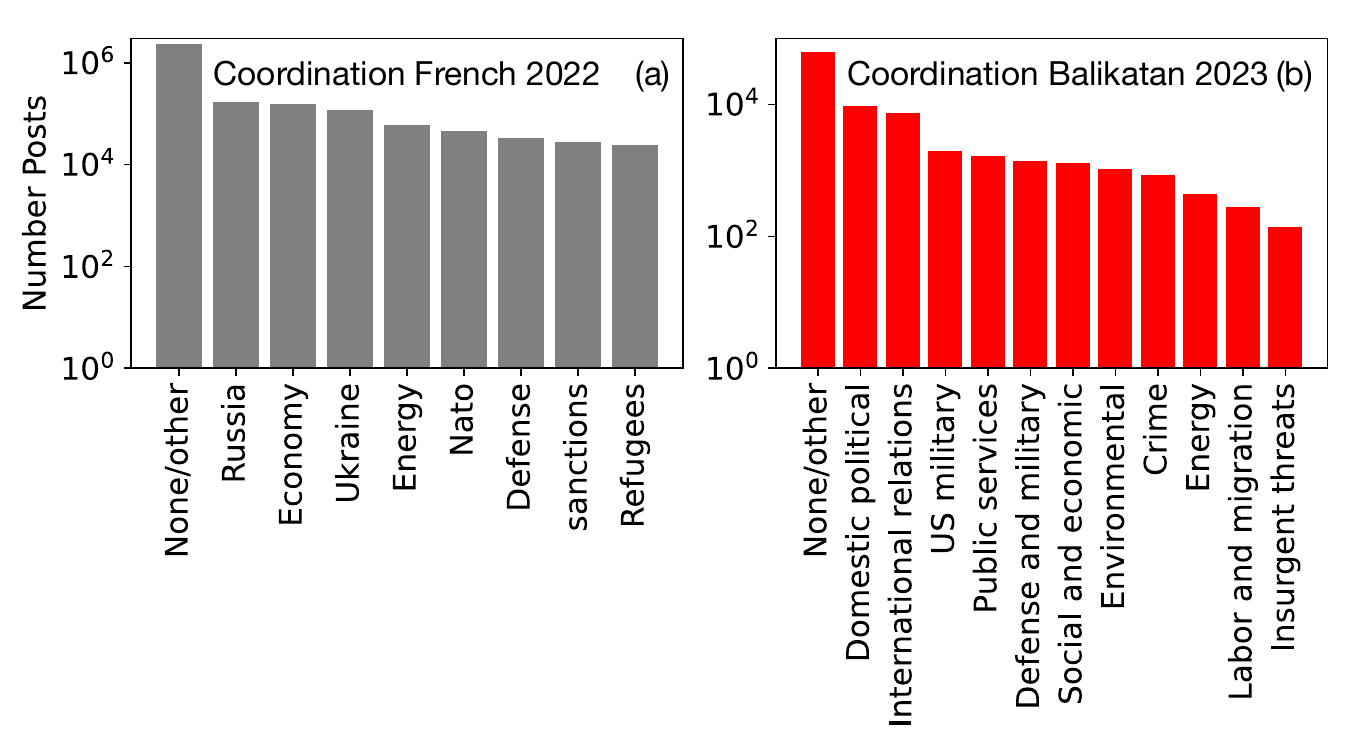}
        \caption{The frequency of concern posts in coordinated accounts.  (a) 2022 French election and (b) 2023 Balikatan U.S.-Philippines military exercises.}
    \label{fig:concern-freq}
\end{figure}

We also analyze the concerns extracted from these coordinated accounts in Fig.~\ref{fig:concern-freq}. This figure shows the frequency of posts related to each concern we capture (note the log-scale of the y-axis); Russia is the most often mentioned concern. Meanwhile, the Balikatan dataset most often discusses domestic political issues and international relations. Compared to ground truth data (Fig~\ref{fig:concern-gtfreq}), domestic political and public service concerns are over-represented.

\begin{figure}
    \centering
    \includegraphics[width=\linewidth]{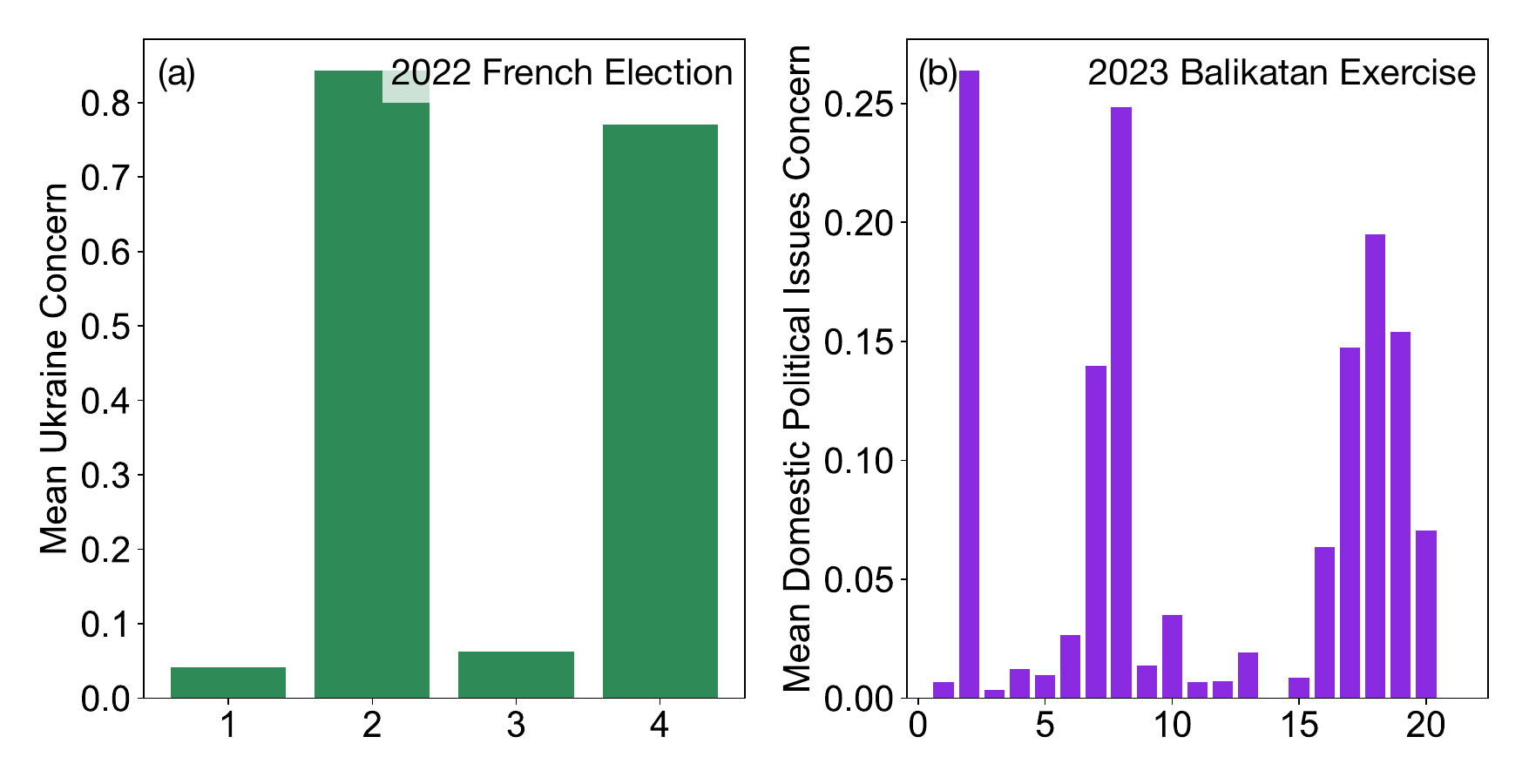}
    \caption{Example concerns for different clusters in the 2022 French election and Balikatan datasets. The x-axis are the cluster IDs, ranked from largest to smallest, where sizes are shown in Fig.~\ref{fig:coord_freq}. (a) Ukraine concern in the 2022 French election and (b) domestic political concern in the 2023 Philippines Balikatan dataset.}
    \label{fig:concern-cluster}
\end{figure}

But these concerns are highly selective in where they appear. We show examples of the percent of posts discussing two example concerns within each coordinated cluster in Fig.~\ref{fig:concern-cluster}. These figures show the overabundance of the Ukraine concern within two clusters of the French election (Fig.~\ref{fig:concern-cluster}a), which, as seen in Fig.~\ref{fig:examples}, corresponds to pro-Ukraine messages. In contrast, Fig.~\ref{fig:concern-cluster}b shows an overabundance of the domestic political concern, especially for the second largest coordinated cluster. This cluster contains an anti-Conservative UK party information campaign (i.e., the posts attack Tories). 

\subsection{Applying LLMs to Coordinated Account Posts}
We will look at two coordinated campaigns as case studies. Due to space limitations we only briefly summarize how they change for various concerns, but we have all results in the link at the top of the paper. We explore two ways that these data are subdivided to better understand information operations: splitting these data by events (e.g., before versus after an election) and splitting these data by the concern the posts share. We qualitatively observe strong differences between what coordinated campaigns discuss before and after events, thus motivating the dynamic nature of our exploration. We also notice that concerns offer more nuance to the claims (e.g., why they are sharing pro-Ukraine concerns) compared to data that is not split by concerns. 

For the 2022 French election, we sampled 20 random posts before the election, between rounds 1 and 2, and after round 2. We specifically focus on accounts sharing the \textit{Ukraine concern} to reduce noise in the captured posts (campaign we study in detail, shown in Fig~\ref{fig:examples} is heavily pro-Ukraine and requests assistance from France).  

When studying the attributes from \cite{martin2019trends}, we notice, however, country attribution is imperfect, where before, during, and after the round 1 and 2 elections, the targeted country is Ukraine rather than Macron or France. The country responsible for the information campaign is either Russia (before the campaign) or France (between rounds 1 and 2, and after round 2). This suggests the LLM confuses attribution (claiming Russia is responsible for the campaign given it is mentioned in the posts). It narrows down the list of potential candidate countries, but the model nonetheless misunderstands the posts. That being said, the political goal is consistent across time (e.g., before round 1, the LLM states): ``Influence international actors to take action against Russian aggression in Ukraine'' which agrees with expectations. 

When studying the BEND framework, we come across surprising utility but also limits of these zero-shot prompts. Namely, when asking for Explain tactic posts, we find the model appears to be spot on: after listing example posts, it says ``These posts provide specific examples of the destruction caused by Russian aggression in Ukraine and call for action to protect Ukrainian civilians. They elaborate on the situation and highlight the urgency of the crisis.''. In contrast, when asking for the Excite tactic prior to the first round, an example output is:

``...2. @EmmanuelMacron NOW!!! RUSSIA FIRES ON NUCLEAR POWER PLANT IN \#UKRAINE!!! STREAM: https://XXXX WAKE UP, WORLD! \#StopPutin \#StopRussia CLOSE THE SKY OVER UKRAINE!!!''

``This tweet uses capital letters and exclamation marks to convey a sense of urgency and excitement in calling for action to stop Russian aggression against Ukraine.'' 

This shows how the LLM mistakes activation for excitement and misunderstands whether positive sentiment exists in the message. These are seen in other examples, such as the Distract tactic, where the LLM states ``These tweets distract from the urgent situation in Ukraine by discussing EU candidate status for Ukraine, calling out a specific individual (Putin), and expressing gratitude for support without any further context related to the conflict. By introducing irrelevant topics, they divert attention away from the main issue at hand which is the ongoing crisis in Ukraine.'' The misunderstanding of distraction may be due to a lack of ``typical'' (non-coordinated) posts listed as counter-examples. In addition, few-shot and Chain-of-Thought prompting might further improve the output quality \cite{zhang2022automatic}.

Finally, when analyzing the framing narrative prompts, we find the LLM is particularly well-suited. The problem is stated  as ``Russian aggression against Ukraine...'' before the round 1 election, and similarly in other time periods. The cause is ``...Russian aggression against Ukraine'' which may be due to a lack of context, but it's also unclear what a human might say. Finally, the remedy is: ``1. Stop Russian aggression against Ukraine. 2. Close the sky over Ukraine. 3. Exclude Russia from the UN Security Council. 4. Ban Russia from SWIFT. 5. Protect the Ukrainian sky. ...'' etc., which is a good summary of the results. Catchphrases and slogans include ``STOP RUSSIAN AGGRESSION AGAINST \#UKRAINE, CLOSE THE SKY OVER UKRAINE, EXCLUDE RUSSIA FROM THE @UN SECURITY COUNCIL, \#StopPutin'' etc. (taken directly from the posts fed into the model), which shows slogans in the posts double as remedies.

Due to limited space, we will briefly mention that later stages, such as after round 2, we see more plausible and concrete remedies as the war progressed such as, ``1. Urging world leaders, particularly Emmanuel Macron, to negotiate a safe passage for the civilians trapped in Mariupol to evacuate and receive humanitarian aid. 2. Providing weapons and support to Ukraine to help defend against Russian aggression and protect the civilians.'' etc., 

For the 2023 Balikatan dataset, we analyze posts from the largest coordinated cluster before and after the Balikatan military exercise. This campaign promotes Pathaan, a Bollywood movie. While this is a coordinated campaign, it lies outside of typical campaigns studied by \cite{martin2019trends} and is more akin to spam. We study the None or Other concern to focus on the most relevant posts for this set of coordinated accounts (which do not discuss otherwise common concerns, such as Crime or US Military). GPT-3.5 states that the country targeted (and the country responsible for the information operation) is consistently India. While the government of India is likely not responsible, the audience and potential coordinators of the information campaign are likely within India due to the genre of the movie. Before April 11, 2023, the political goal is stated as ``Promoting the Bollywood movie `Pathaan''' which hints at another utility of the GPT-3.5 outputs: with one prompt, the political goal, we can immediately determine that those accounts were not related to ``important'' coordinated accounts (i.e., those about Balikatan or targeting democracies). We can also easily understand the underlying role of these spam accounts because after April 28, 2023, they were pushing messages about ``Adipurush'', a separate Indian movie. Moreover, there appears little overlap in the directors or producers to each film allowing us to speculate that this set of X accounts is run by a group that offers their services to a number of different organizations. We also find the LLM extracts relevant BEND tactics, such as posts promoting excitement (e.g., ```...book now!''). Finally, framing theory prompts reveal: ``The cause of the problem mentioned in these tweets is the high demand and anticipation for the movie `Pathaan.'...'' and ``The remedy to the problem mentioned in the tweets is for individuals to book their tickets in advance for the movie `Pathaan''' which shows the accounts are trying to drive ticket sales. 

\section{Discussion and Conclusions}

To summarize, we worked with domain experts to curate a list of concerns associated with two distinct datasets, a 2022 French Election dataset and a 2023 Balikatan dataset. We then developed a framework to detect these concerns using an instruction-tuned Llama-2 LLM. We find that the Llama-based concern detection model achieves high accuracy, making it suitable for our analysis. We then applied GPT-3.5 to annotate coordinated campaigns that contained each of these concerns. Namely, we use GPT-3.5 to extract the goals, tactics, and frames that are based on information operation descriptions from an information operations dataset \cite{martin2019trends}, the BEND framework \cite{carley2020}, and Framing Theory \cite{chong2007framing}. GPT-3.5 was first evaluated against 126 known campaigns and achieved reasonable accuracy. We then detected coordinated campaigns within the two datasets via a standard hashtag-based method \cite{burghardt2023socio}, and annotated campaigns via the GPT-3.5 LLM. These posts were first filtered by the concerns they shared and when these posts were made to help distill the main properties of dynamic information operations. When GPT-3.5 was applied to these campaigns, we found the LLM was effective at extracting the goals, tactics, and narrative frames, although it made notable errors.  While GPT-3.5 is not always correct, the results are still informative for large sets of coordinated accounts. LLMs appear likely to help stake-holders find and understand the behavior of various information operations at scale and find common patterns across information operations, which could otherwise be missed from purely manual annotation. While LLMs still require a human in the loop to interpret these results, we find in practice that verifying LLM annotations is much faster than humans creating their own annotations, much like how LLMs can speed up letter-writing, coding, and other tasks that still require a human in the loop.

\subsection{Limitations}
There are a few important limitations of our work. First, LLMs can hallucinate. Detecting hallucinations is hard to automate at present, therefore we need a human in the loop to verify any output. Efforts to detect or address hallucinations, especially to annotate coordinated campaigns, needs to be analyzed. Moreover, the accuracy of a LLM depends strongly on carefully constructed prompts. The degree to which semantically equivalent prompts may affect the output of the model needs to be further explored. 

 In addition, the low performance of the concern detection model on some concerns such as \emph{refugees} and \emph{defense} suggests that Llama or GPT-4 (whose training data we rely on) may fail to understand the nuance of these concerns. Finally, the reproducibility of results depends strongly on the model version and the stochasticity (e.g., temperature) of the output. These are challenges for any replication study, especially for LLMs that are regularly updated.

\subsection{Ethical Considerations}
Our work utilizes public data on information campaigns and anonymized social media data, therefore we believe the harm of this research on any human subjects is minimal. The potential harm of this research, however, is that we might provide ways for information campaigns to better hide their underlying goals, or to scrutinize other coordinated campaigns to improve their influence operation. We believe, however, that these potential harms are minimal as the information operations already try to hide their behavior to avoid their accounts being suspended, and the campaign operators could already scrutinize known campaigns that have been publicly reported. In addition, there might be costs to misclassifying accounts as coordinated or LLMs misunderstanding a coordinated campaign by flagging or misunderstanding text from authentic accounts. Care must therefore be taken to assume individual accounts are authentic until proven otherwise.

\subsection{Future work}
There are a range of ways to improve the concern detection and LLM annotation models. 

The concern detection model could be improved with additional human-annotated training data. This can be applied as gold standard instruction-tuning training data, for GPT-4 few-shot learning prompts in order to improve the synthetic training data, or to up-sample concerns with very few annotated post (such as public services or energy concerns in the Balikatan dataset) in order to better validate any future model. Newer models, such as Llama-3 \cite{huang2024good} or larger models, such as Llama-2-70B could also improve the concern detection accuracy and should be explored in the future. 

There are also a number of ways to improve coordination campaign annotation. First, we can provide few-shot or chain-of-thought \cite{zhang2022automatic} prompting to improve the accuracy of the output. In addition, a collection of domain experts should independently verify whether the model can, for example, approximately reconstruct the major features of known coordinated campaigns \cite{martin2019trends}, as well as unknown campaigns. This will improve the validation of our observations.  Finally, we can use larger models, such as GPT-4 \cite{achiam2023gpt}, to improve the quality of the annotations.

%\bibliography{references}

\section{Paper Checklist}

\begin{enumerate}

\item For most authors...
\begin{enumerate}
 \item  Would answering this research question advance science without violating social contracts, such as violating privacy norms, perpetuating unfair profiling, exacerbating the socio-economic divide, or implying disrespect to societies or cultures?\\
    Yes, see Ethical considerations.
  \item Do your main claims in the abstract and introduction accurately reflect the paper's contributions and scope?\\
    Yes, for example, we made sure to state both the promise and limitations of our results throughout the paper.
   \item Do you clarify how the proposed methodological approach is appropriate for the claims made? \\
    Yes, we were careful to state the purpose of our methods and its limitations.
   \item Do you clarify what are possible artifacts in the data used, given population-specific distributions?\\
    Yes, see Ethical Considerations.
  \item Did you describe the limitations of your work?\\
    Yes, see Limitations within the Discussion and Conclusions.
  \item Did you discuss any potential negative societal impacts of your work?\\
    Yes, see Ethical Considerations.
      \item Did you discuss any potential misuse of your work?\\
   Yes, see Ethical Considerations.
    \item Did you describe steps taken to prevent or mitigate potential negative outcomes of the research, such as data and model documentation, data anonymization, responsible release, access control, and the reproducibility of findings?\\
   Yes, see Ethical Considerations.
  \item Have you read the ethics review guidelines and ensured that your paper conforms to them?\\
    Yes.
\end{enumerate}

\item Additionally, if your study involves hypotheses testing...
\begin{enumerate}
  \item Did you clearly state the assumptions underlying all theoretical results?\\
    NA.
  \item Have you provided justifications for all theoretical results?\\
    NA.
  \item Did you discuss competing hypotheses or theories that might challenge or complement your theoretical results?\\
    NA.
  \item Have you considered alternative mechanisms or explanations that might account for the same outcomes observed in your study?\\
    NA.
  \item Did you address potential biases or limitations in your theoretical framework?\\
    NA.
  \item Have you related your theoretical results to the existing literature in social science?\\
    NA.
  \item Did you discuss the implications of your theoretical results for policy, practice, or further research in the social science domain?\\
    NA.
\end{enumerate}

\item Additionally, if you are including theoretical proofs...
\begin{enumerate}
  \item Did you state the full set of assumptions of all theoretical results?\\
    NA.
	\item Did you include complete proofs of all theoretical results?\\
    NA.
\end{enumerate}

\item Additionally, if you ran machine learning experiments...
\begin{enumerate}
  \item Did you include the code, data, and instructions needed to reproduce the main experimental results (either in the supplemental material or as a URL)?\\
    Yes, see code link in Introduction.
  \item Did you specify all the training details (e.g., data splits, hyperparameters, how they were chosen)?\\
    Yes, see Methods and code.
     \item Did you report error bars (e.g., with respect to the random seed after running experiments multiple times)?\\
    Yes, see Tables 1 and 2.
	\item Did you include the total amount of compute and the type of resources used (e.g., type of GPUs, internal cluster, or cloud provider)?\\
    Yes, see Methods.
     \item Do you justify how the proposed evaluation is sufficient and appropriate to the claims made? \\
    Yes, see Results and Discussion.
     \item Do you discuss what is ``the cost`` of misclassification and fault (in)tolerance?\\
    Yes, see Ethical Considerations.
  
\end{enumerate}

\item Additionally, if you are using existing assets (e.g., code, data, models) or curating/releasing new assets, \textbf{without compromising anonymity}...
\begin{enumerate}
  \item If your work uses existing assets, did you cite the creators?\\
    NA.
  \item Did you mention the license of the assets?\\
    NA.
  \item Did you include any new assets in the supplemental material or as a URL?\\
    Yes, see 
  \item Did you discuss whether and how consent was obtained from people whose data you're using/curating?\\
    Yes, see Data subsection with in Methods.
  \item Did you discuss whether the data you are using/curating contains personally identifiable information or offensive content?\\
  Yes, see Data subsection with in Methods.
\item If you are curating or releasing new datasets, did you discuss how you intend to make your datasets FAIR (see \citet{fair})?\\
NA.
\item If you are curating or releasing new datasets, did you create a Datasheet for the Dataset (see \citet{gebru2021datasheets})? \\
NA.
\end{enumerate}

\item Additionally, if you used crowdsourcing or conducted research with human subjects, \textbf{without compromising anonymity}...
\begin{enumerate}
  \item Did you include the full text of instructions given to participants and screenshots?\\
  NA.
  \item Did you describe any potential participant risks, with mentions of Institutional Review Board (IRB) approvals?\\
    Yes, see Methods for IRB approval and Ethical considerations for potential participant risks.
  \item Did you include the estimated hourly wage paid to participants and the total amount spent on participant compensation?\\
   NA.
   \item Did you discuss how data is stored, shared, and deidentified?\\
   Yes.
\end{enumerate}

\end{enumerate}

\section{Appendix}

\subsection{Prompting Template}
\label{sec:prompt_temp}

\noindent \texttt{You are using an AI model specialized in classifying tweets into relevant [event topic] concern (wedge issues) categories. The available concern categories are [concern categories]. Your task is to analyze the following tweet and categorize the following tweet into one or more of these concern categories. If the tweet does not align with any of these categories, label it as \#NoneOther. Please respond with only the relevant category labels.}\\

\noindent \texttt{Tweet: [tweet]}\\

\subsection{Instruction Template}
\label{sec:inst_temp}

\noindent \texttt{Instruction: You are using an AI model specialized in classifying tweets into relevant [event topic] concern (wedge issues) categories. The available concern categories are [concern categories]. Your task is to analyze the following tweet and categorize the following tweet into one or more of these concern categories. If the tweet does not align with any of these categories, label it as \#NoneOther. Please respond with only the relevant category labels.}\\

\noindent \texttt{Input: [tweet]}\\

\noindent \texttt{output: [labels]}\\

\subsection{LLM X post generation examples}
\begin{table*}[]
    \centering
    \begin{tabular}{p{6cm}|p{10cm}}
         Political goal& LLM-generated post \\\hline\hline
         Polarize Australian politics&Why is \#Australia letting in so many Muslim refugees when they bring nothing but trouble? \#StopIslam \\\hline
         Discredit Sebastian Kurz after the 2017 Presidential elections in Austria&BREAKING: Shocking revelations have emerged about Sebastian Kurz's connections to Islamic immigration! \#DiscreditKurz \#Austria \\\hline
         Undermine Belarusian government & Just heard that Russia offered Belarus a sweet deal to join forces. Maybe it's time for change \#BelarusStrong\\\hline
    \end{tabular}
    \caption{Political goals in various information operations \cite{martin2019trends} and LLM-generated posts for these campaigns.}
    \label{tab:LLm_ex}
\end{table*}
\end{document}